# An Innovative Word Encoding Method For Text Classification Using Convolutional Neural Network

**Amr Adel Helmy**
College of Computing and Information Technology.
Arab Academy for Science, Technology and Maritime Transport.
Cairo, Egypt
aaheg@msn.com

**Yasser M.K. Omar**
College of Computing and Information Technology.
Arab Academy for Science, Technology and Maritime Transport.
Cairo, Egypt
dr_yaser_omar@yahoo.com

**Rania Hodhod**
Faculty of Computer and Information Sciences
Ain Shams University,
Cairo, Egypt
rhodhod @cis.asu.edu.eg

*Abstract*—Text classification plays a vital role today especially with the intensive use of social networking media. Recently, different architectures of convolutional neural networks have been used for text classification in which one-hot vector, and word embedding methods are commonly used. This paper presents a new language independent word encoding method for text classification. The proposed model converts raw text data to low-level feature dimension with minimal or no preprocessing steps by using a new approach called binary unique number of word "BUNOW." BUNOW allows each unique word to have an integer ID in a dictionary that is represented as a k-dimensional vector of its binary equivalent. The output vector of this encoding is fed into a convolutional neural network (CNN) model for classification. Moreover, the proposed model reduces the neural network parameters, allows faster computation with few network layers, where a word is atomic representation the document as in word level, and decrease memory consumption for character level representation. The provided CNN model is able to work with other languages or multi-lingual text without the need for any changes in the encoding method. The model outperforms the character level and very deep character level CNNs models in terms of accuracy, network parameters, and memory consumption; the results show total classification accuracy 91.99% and error 8.01% using AG's News dataset compared to the state of art methods that have total classification accuracy 91.45% and error 8.55%, in addition to the reduction in input feature vector and neural network parameters by 62% and 34%, respectively.

## I. INTRODUCTION

Text classification uses natural language processing (NLP) to classify text data to its predefined categories. Machine Learning (ML) techniques are intensively used for this task because of their ability to automatically recognize the different complex patterns and distinguish between them.

Text data characterized by very high dimensionality that can cause a phenomenon is called curse of dimensionality [1]. This makes it unsuitable for training using machine learning techniques without applying feature selection methodologies in order to extract meaningful features that can help to reduce the dimensionality. Accordingly, feature extraction and efficient representation of text become important factors in improving the accuracy of classification. Many text representation methods have been used, such as bag-of-words (BOWs) [2], Term Frequency, and Inverse Document Frequency (TF-IDF) [3], Latent Semantic Indexing (LSI) [4] and others in order to select the appropriate features to be used by classifiers like Naive Bayes (NB), k-nearest neighbors (KNN), and support vector machine (SVM).

One of the critical questions posed by researchers in automatic text classification is how to choose the best feature vector for traditional machine learning classifiers? The challenge lies in the fact that there is no single feature engineering technique that can work for all text classification and learning tasks. In other words, there is no best practice as feature engineering has its drawbacks due to loss of information, and need manually fine-tune the data, and requiring prior familiarity with the language.

Recently, convolutional neural networks (CNNs) are well suited for text classification as it outperforms other models, such as bag-of-words (BOWs). CNNs proved its ability to learn automatically from scratch using character-level representations of text irrespective of the language used and without prior knowledge of language words, syntax, grammar and semantic similarities [5]. CNNs allow high level understanding, provided the availability of sufficient data.

The use of word embedding (Word2Vec) technique with CNNs has attracted considerable attention from many researchers over the traditional models like One-hot vector and BOW because of its ability to reduce memory requirements and training time, in addition to its effect on performance. In the case of BOW, each word is represented as one-hot vector with dimensions equal to the words vocabulary size (N); each 1 digit is placed in the correspondent position of that word in a 1-N vector, all other positions are filled with 0 digits. Since natural languages are characterized by huge vocabulary size, it is common to use word frequency statistics or relevance metrics to determine the most frequent words that are representative of the texts and exclude the rare ones. This helps to control the vocabulary size that directly affects the computational and memory requirements.

In the Word2Vec approach, each word is projected into an embedding metric of fixed size that represents its co-occurrence in a text corpus; Mikolov et al. used two widely architectures model, CBOW model and the skip-gram model to compute continuous vector representations of words from very large data sets [6]. Both models use dense vectorization to represent the word vectors. Their results provided state-of-the-art for measuring syntactic and semantic word similarities.

This paper proposes a new language independent word representation technique for feeding text data into a single hybrid character-word CNN classification model. The model achieved highest accuracy result compared to the state of art of CNN models. The proposed model takes advantage of character level in term of less preprocessing on raw data, less features vector dimension, less neural network parameters, in addition to word level in terms of using fewer convolutional layers, where a word is atomic representation the document.

The remainder of the paper is organized as follows. Related work is presented in the next section. The proposed model architecture is described in detail in Section 3. Results compared to the state of art in text classification and experiment methodology are detailed in Section 4. Finally, conclusions and future work are presented in Section 5.

## II. RELATED WORKS

Text classification is one of the most important tasks in Natural Language Processing (NLP). Traditionally, the input features vector of a given document is represented as bag-of-words or n-grams where their frequency–inverse document frequency (TF-IDF) serves as the input for a subsequent linear classifier. The most popular classifiers used for text classification are support vector machines [7], naive bayes [8], and LIBLINEAR it supports logistic regression, and linear support vector machines [9].

Recently, deep learning methods, such as CNNs [5, 10, 11] are used for feature extraction and classification of text while bag-of-words (BOWs) [12] and Word2Vec [6] are used to represent textual information. The outputs from applying these methods then serve as input to CNNs. Researchers in [10] proposed a shallow neural network with one convolutional layer (using multiple filter widths and feature maps), followed by a max over time pooling layer and one fully connected layer with dropout and softmax output. This convolutional layer works on top of initialized input word vectors from text data; each word is of k-dimensional size and is obtained from its correspondence pre-trained Word2Vec on 100 billion words of Google News [6]. Other researchers followed the same methodology of using CNN layers, in addition to introducing a max pooling layers with dynamic k size [13]. This assists in detecting the highly important k features in a text, regardless of their position, while taken in consideration their relative order (the network layer position and text length determine the value of k).

Many researchers have observed that it is not necessary to use the one-hot vector representation of word (word-level) with a deep neural network and can be used character or even sub-word level as an alternative. Researchers introduced a use of character sequence as a substitution to the word one-hot vector [14, 15] because of its low representation vector as character depends only on the number of definite characters exist in the language used compared to huge vocabulary size associated with word representation. Character concept also used instead of word for dependency parsing in [16].

Others noted that the word representation could be unsuitable for social media like Twitter, where tokens are usually a challenge due to the existence of slang, elongated words, contiguous sequences of exclamation marks, abbreviations, and hash tags [17]. Therefore, they introduced a characters-level CNN, which automatically learns the words and notions of sentences from scratch and without pre-processing, or even tokenization. The most relevant work on character-level CNN for text classification was proposed by [5] and [11] - 6 CNN layers and 29 CNN layers, respectively.

More work was done on using CNN to train classifiers by representing text in an image like fashion where each character has atomic representation [5]. This enabled a deep CNN to classify text with high-level concepts. In this work, each English alphabetic character inside the document of length l (l is the length sequences of character in the document) is encoded in a one-hot vector (one-of-m) with a value of one in the position of this character inside the vector, and zero values for the other, where size m is equivalent to alphabet vocabulary size. Using this encoding method, data is fed into a CNN model that consists of six 1-D temporal layers and three fully connected layers. Kernels size of three and seven are used in addition to simplest max-pooling layers. Features are extracted from small, overlapping windows of the input sequence of each layer and pools over small, non-overlapping windows by taking the maximum activations in the window. This method performs entirely at the character level and is able to learn from scratch with minimal processing and without prior knowledge of the language structure. This work shows that character-level CNN is an effective method.

The work presented in [18] introduced CNN architecture with recurrent layers where a one-hot sequence input created from character vocabulary of ninety-six case sensitive characters, digit numbers, punctuation, and spaces, is converted to a vector dense of size 8 using an embedding layer. The output of this layer is then fed into multiple convolutional layers of kernel size three or five depending on the depth and max-pooling layer of size two with rectified linear unit (ReLU) activation function to get a shorter feature vector [19]. This feature vector is then fed into a single recurrent layer with bidirectional Long-short term memory (LSTM) where it concatenates the last hidden state of both directions forming a fixed-dimensional vector. Finally, this vector is fed into the classification layer to compute the probabilities of each category. The researchers in this work applied dropout after the last convolutional layer to avoid deep neural networks over-fitting, and recurrent layer. This hybrid model was able to capture sub-word information from a character sequence in a document, and achieve comparable performances compared to [5].

Work done in [11] proposed a new character level CNN with very deep CNN (up to 49 layers) inspired by VGG and ResNets like architecture philosophy. The method used in this work starts with applying a lookup table to create a 2D dense vector where each character in fixed document length of 1024 was converted to a vector size of 16. The produced vector is then fed into 64 feature map convolutional layer of kernel size three, followed multi-convolutional a stack blocks. Each convolutional block consists of multi-temporal convolutional layers with different feature maps that depend on the layer depth and fixed kernel size of three [20]. This layer is followed by a temporal batch norm layer and a ReLU activation function. Each convolutional block is followed by a max-pooling layer of kernel size 3 and strides 2, and then double the size of convolutional feature maps of the next block. In the last block, they selected k most important features 512 × k and transformed them to a vector, which is then fed into a three fully connected layer with ReLU activation function and softmax outputs. This architecture achieved high accuracy result while using less network parameters compared to character level CNN.

The most commonly used method for training CNN for text classification depends on applying a Word2Vec [6] or Global vectors [21] or embedded layer on word or character input feature vector to extracts a low representing feature vector to avoid the curse of dimension, and enhance the learning process especially with word level or by applying 1-of-m encoding to learn from scratch, especially with character level [5]. All these methods produce a large input feature vector in addition to its application to a specific language and if other language is used, many variables need to be re-engineered.

## III. PROPOSED MODEL

This paper proposes a CNN model that uses an innovative method called Binary Unique Number of Word Method (BUNOW) to encode text (full description for BUNOW is provided in Section (III.A). The output of encoding is then fed into the CNN (CNN architecture is detailed in Section III.B). The model allows the creation of a 2D input feature vector where each word exists in a document is represented by a fixed k-dimensional vector. The vector is then inserted into a document in a subsequent position relative to its original position. The goal is to increase the accuracy while decreasing the neural network parameters to reduce memory usage and computational consumption.

### A. BINARY UNIQUE NUMBER OF WORD METHOD (BUNOW)

The rationale behind the BUNOW method is to convert the unique serial integer ID ($ID_{wi}$) given to each unique word ($w_i$) in the training corpus (T) to its binary equivalent number ($B_{wi}$), and represent it as a fixed k-dimensional vector where its dimension equal to ($2^k$ = Total $ID_{wi}$ in T). Figure 1 shows the flow of the BUNOW method in which the following steps are applied:

1) Create vocabulary of unique words from training dataset.

2) Assign unique serial integer ID for each word.

3) Convert the serial integer ID for each word to its equivalent binary value, and represent it in a fixed dimensional vector (b) of size k, where ($2^k$ = vocabulary size).

4) Set the maximum document length ($L_{Wmax}$) based on the document with maximum number of words ($W_{max}$) in the training dataset.

The input feature vector ($I_V$) is created by concatenating a binary vector of each word ($b_{wi}$) that exists in a document (D), such that:

$$I_V D = b_{w1} + b_{w2} + b_{w3} + b_{w4} + ... + b_{Wmax} \quad (1)$$

Where (+) is the concatenation operator.

Any word that exceeds the document length ($L_{Wmax}$) is ignored, and any document less than ($L_{Wmax}$) are padded with all-zero vectors. In the testing phase, instead of using zeroes for an unseen word, we gave each unique unseen word an equivalent binary value equal to its subsequent integer ID in vocabulary (V) and then updates V under the following constrain: the representation of the binary value of an unseen word would not exceed the initialized fixed dimensional vector (b) of the word, if exceeded then a vector of zeros for this word will be used.

### B. CNN ARCHITECTURE

The provided model in this paper is partially inspired by Alex-Net [22], a large deep CNN model that competed in 2012 challenge (Image-Net Large Scale Visual Recognition) and was classified as a top-5 winner achieving an accuracy rate of 84.7% compared to 73.8% achieved by the second-best entry. Alex-Net architecture is characterized by using a stack of convolutional layers with max pooling, and ReLU activation function, three fully connected layers with dropout 0.5, and softmax classifier in the last fully connected layer [23]. A similar architecture was used by [5] for text classification where the researchers used six 1D temporal convolutional layers of kernels size 3, and 7 with convolutional feature maps of 256 for small feature, while using 1024 for large feature followed by simple Max-pooling layers, and three fully connected layers.

The architecture of the proposed CNN is summarized in Figure 2. It consists of eight deep spatial convolutional layers, which simply computes a 2-D convolution and three fully connected layers. The architecture is divided into three blocks; the first block consists of four convolutional layer with feature maps 200, stride (1, 1), and ReLU activation function for each. Vertical convolutional is applied over input vector while ignoring horizontal convolution by using kernel size ((one, 7), (one, 7), (one, 3), (one, 3)) respectively with valid padding to learning the structure of each word separately from its binary k-dimensional vector. This allows the network to understand the words and be able to distinguish between them.

The second block consists of four convolutional layer with stride (1, 1), ReLU activation function for each layer, and feature maps (150, 100, 50, 25), respectively. Horizontal convolutional is then applied over output feature maps of the previous block by using kernel size ((three, 1), (three, 1), (seven, 1), (seven, 1)) respectively with valid padding in order to let the network understand the semantic and the relation between words inside the document and distinguish between them.

Finally, three fully connected layers of 2048, 1024, and 4 are used after the last convolutional layer in which the last fully connected layer connect to softmax classifier to classify the document based on four class category.

In order to regularize the network and prevent over-fitting, a five dropout of 0.5 is used after the 4[th], 6[th], 8[th], 9[th], and 10[th] layer, respectively [24].

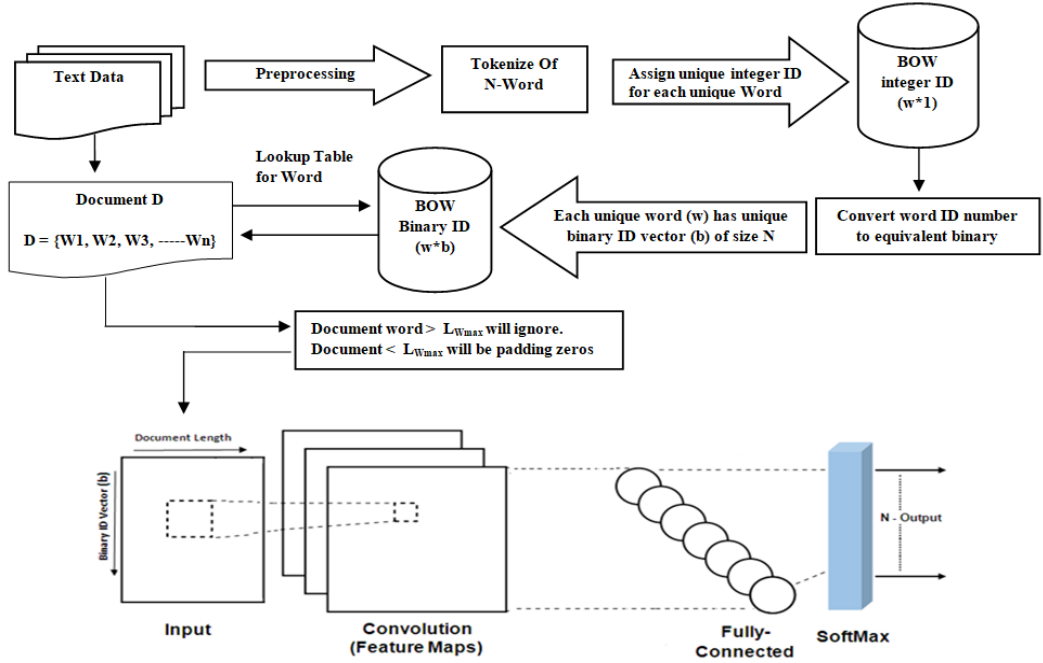

FIGURE 1 Binary Unique Number of Word (BUNOW)

A max-pooling layer is commonly used to reduce a feature map size of output convolutional layer by selecting the maximum feature value before connecting it to another convolutional layer or a fully connected layer. This way allows for fewer weight parameters, which directly affects memory and computational consumption. It was found that it is not useful to use the pooling layer with low-level convolutional feature map of size 165 each as this leads to loss of information. The alternative was to reduce the convolutional feature maps number from 200 to 25 in the last convolutional layer. This helps to achieve high accuracy without loss of information compared to the current state of the art.

## IV. EXPERIMENT SCENARIO & RESULTS

### A. DATASET

The dataset used is from AG's News that can be obtained from news articles on the web [4]. It approximately contains of 496K articles extracted from more than 2000 news sources. The selected categories from this corpus are the four largest ones (Sports, Sci/Tech, Business, and World). The title and description fields are used from these categories to build this dataset. From each category, 30,000 samples were randomly chosen for training and 1,900 for testing [5]. Samples size is about 120,000 for training and 7,600 for testing. The dataset statistic is shown below in Table 1.

TABLE 1  AG'S NEWS STATISTIC

| Class # | Language | Training # | Testing # | Unique Words # | Max Doc Length # |
|---|---|---|---|---|---|
| 4 | English | 120,000 | 7,600 | 70,396 | 181 |

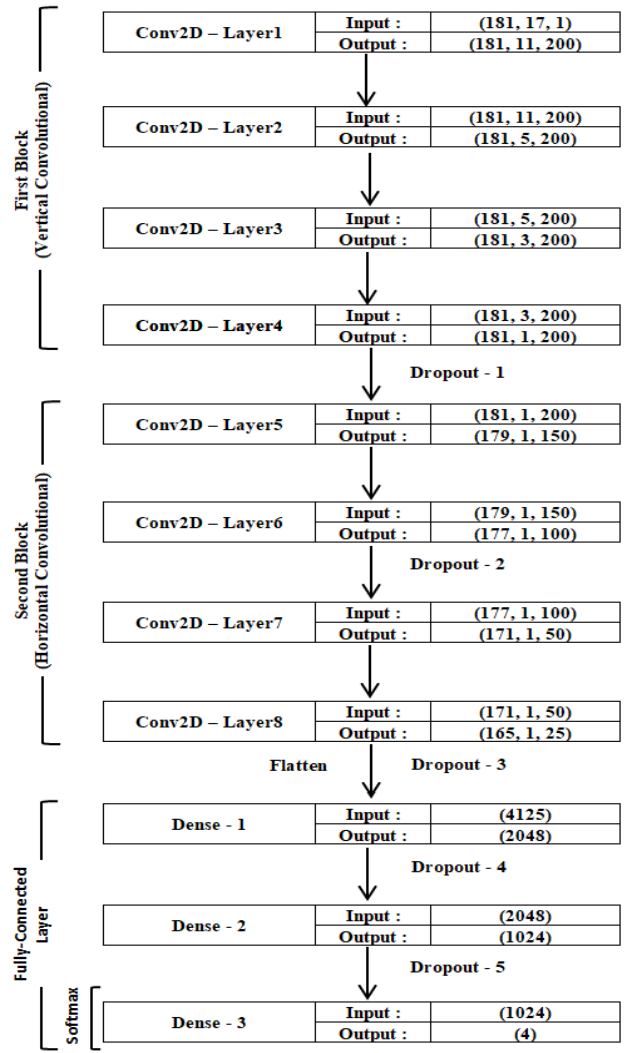

FIGURE 2 Visualization of CNN Architecture

### B. MODEL SETTINGS

The following setting has been taken under consideration:

1) Processing is performed on word-level which is the atomic representation of a text document.
2) The vocabulary size consists of 70396 unique words after preprocessing by removing (! " # $ % & ( ) * + , - . / : ; < = > ? @ [ \\ ] ^ _ ` { | } ~ \t \n), and convert all words to lower case.
3) Post padded technique used to ensure all input document has same length of 181.
4) For each word, a BUNOW vector length is 17.
5) Training is performed with Adam optimizer, and Categorical Cross-Entropy loss function, using a batch size of size 120, learning rate of 0.0005.
6) Neural network weights are initialized using Xavier Uniform Initializer [25], and bias initialize with zeroes.
7) One epoch took 4.3 minute. It took 60 epochs to converge where data shuffle on each epoch.
8) Implementation is done using python and deep learning library (Keras) [26] with Tensorflow backend [27].
9) Single GeForce GTX 1060 Max-Q with GPU memory 6G DDR5 (laptop version)

### C. RESULTS

Our results are shown in Table 2. Test Error percentage was measured for each class along with overall Test Error percentage that was computed by the following equation

$$Test\ Error\ \% = \frac{Number\ of\ incorrect\ results}{Number\ of\ total\ results} \times 100 \qquad (2)$$

TABLE 2  OUR PROPOSED MODEL TEST ERROR %

| AG's News dataset | | | | |
|---|---|---|---|---|
| **Class** | **World** | **Sports** | **Business** | **Sci/Tech** |
| **Test Error %** | 9.53 | 2.63 | 11.84 | 8.05 |
| **Overall Test Error %** | 8.01 | | | |

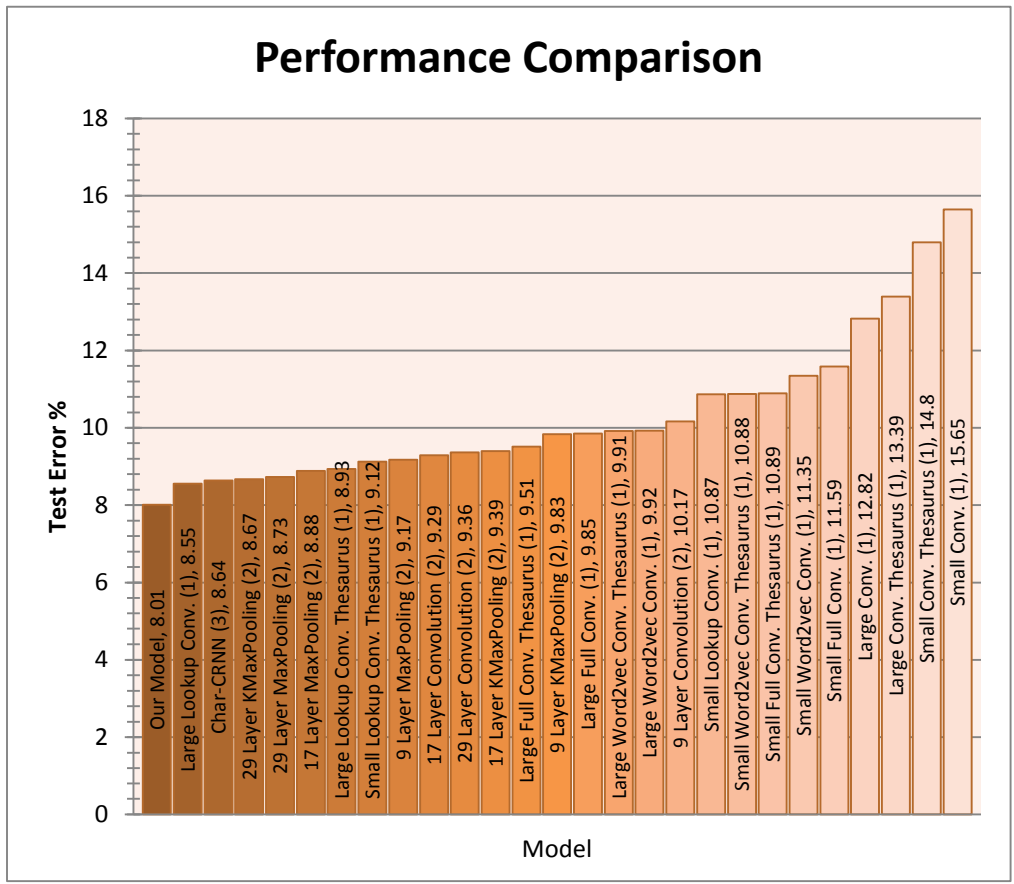

FIGURE 3. Visually compares the performance, where (1) representing to (Zhang et al., 2015), and (2) representing (Conneau et al., 2017), and (3) representing (Xiao and Cho, 2016)

Figure 3 visually compares the performance of the proposed model with the other CNN models results achieved from 16 experiment introduced in [5], and 9 experiment models introduced in [11], in addition to convolution recurrent network (char-CRNN) introduced in [18] using AG's News dataset. The performance of our approach compared to other approaches is shown in Table 3.

The proposed model achieved a better overall test error rate of 8.01% compared to 8.55% achieved by the best one in terms of accuracy. Note: a comparison based on individual classes was not possible as no information related to individual classes was provided in [5, 11, and 18]. More importantly, the input feature vector size and number of neural network parameters in the proposed model are less than the best one by almost 62% and 34%, respectively.

TABLE 3    COMPARISON RESULT SUMMARY

An embedding layer transforms each character to a k-dimensional vector space in order to reduce the one-hot vector size of character before feeding into CNN. Parameter# is approximated.

| Best of | Test Error % | Embedded Layer Size | CNN Input Vector Size | Parameter # in Million |
|---|---|---|---|---|
| Zhang et al., 2015 | 8.55 | N/A | 1014 x 70 | 95 |
| Xiao et al., 2016 | 8.64 | 96 x 8 | 1014 x 8 | 20 |
| Conneau et al., 2017 | 8.67 | 69 x 16 | 1014 x 16 | 17.2 |
| Proposed | 8.01 | N/A | 181 x 17 | 11.3 |

## V. CONCLUSIONS AND FUTURE WORK

This paper presents a new simple encoding method "BUNOW" used for feeding the raw text data into spatial CNN architecture instead of commonly used methods like one hot vector or word representation (i.e. word2vec) with temporal CNN architecture. The main core idea depends on representing each word as a fixed k-dimensional binary vector equivalent to the unique integer ID of this word inside the dictionary. Our model is language independent with small input feature vector, which allows for less number of neural network parameters. Compared to 26 research experiments used the same dataset, the proposed model achieved 8.01 % test error, which is lower than any of the recorded results achieved by the current CNN models. Despite the promising results achieved by the proposed model, future work involves the use of the model to classify multi-lingual text and measure its effectiveness in large-scale datasets.